\documentclass[
]{ceurart}

\usepackage{listings}
\usepackage{svg}
\usepackage{adjustbox}
\begin{document}

\copyrightyear{2024}
\copyrightclause{Copyright for this paper by its authors.
  Use permitted under Creative Commons License Attribution 4.0
  International (CC BY 4.0).}

\conference{First international workshop on Generative AI for Learning Analytics (GenAI-LA),in the 14th International Conference on Learning Analytics and Knowledge (LAK24)}

\title{Supporting Student Decisions on Learning Recommendations: An LLM-Based Chatbot with Knowledge Graph Contextualization for Conversational Explainability and Mentoring}

%%\tnotemark[1]
%%\tnotetext[1]{You can use this document as the template for preparing your
%%  publication. We recommend using the latest version of the ceurart style.}

\author[1]{Hasan Abu-Rasheed}[%
email=hasan.abu.rasheed@uni-siegen.de
]
\cormark[1]
%%\fnmark[1]
\address[1]{University of Siegen, Siegen, Germany}

\author[1]{Mohamad Hussam Abdulsalam}[
]
%%\fnmark[1]

\author[1]{Christian Weber}[
]
%%\fnmark[1]
\author[1]{Madjid Fathi}[
]
%%\fnmark[1]

\cortext[1]{Corresponding author.}
%%\fntext[1]{These authors contributed equally.}

\begin{abstract}
  Student commitment towards a learning recommendation is not separable from their understanding of the reasons it was recommended to them; and their ability to modify it based on that understanding. Among explainability approaches, chatbots offer the potential to engage the student in a conversation, similar to a discussion with a peer or a mentor. The capabilities of chatbots, however, are still not sufficient to replace a human mentor, despite the advancements of generative AI (GenAI) and large language models (LLM). Therefore, we propose an approach to utilize chatbots as mediators of the conversation and sources of limited and controlled generation of explanations, to harvest the potential of LLMs while reducing their potential risks at the same time. The proposed LLM-based chatbot supports students in understanding learning-paths recommendations. We use a knowledge graph (KG) as a human-curated source of information, to regulate the LLM’s output through defining its prompt’s context. A group chat approach is developed to connect students with human mentors, either on demand or in cases that exceed the chatbot’s pre-defined tasks. We evaluate the chatbot with a user study, to provide a proof-of-concept and highlight the potential requirements and limitations of utilizing chatbots in conversational explainability.
\end{abstract}

\begin{keywords}
  Explainable AI (XAI) \sep
  Decision support \sep
  Recommender systems \sep
  Generative AI (GenAI) \sep
  Large language models (LLM) \sep
  Chatbot \sep
  Conversational explanations \sep
  Recommender systems \sep
  ChatGPT \sep
  GPT-4
\end{keywords}

\maketitle

\section{Introduction}

Supporting students with continuous assistance, guidance, and feedback during the learning process is an important concept that has strong foundations in social constructivism and Vygotsky’s Zone of Proximal Development \cite{vygotsky_mind_1978}, as well as scaffolding theory \cite{wood_role_1976}. The availability of an experienced peer or a mentor is, however, one of the main challenges in online learning, especially with the vast amount of online learning resources. Therefore, the research on recommendation systems (RS), explainability and automated feedback systems has been a growing interest in the domain of technology enhanced learning (TEL) and decision support in education.

Presenting learners with explanations and additional information about the educational recommendations they receive has shown promising results in improving their acceptance of these suggestions \cite{ooge_explaining_2022}. Learning-recommendation explainability serves a dual purpose: clarifying the reasons behind specific content suggestions, and empowering the learner’s ability to make an informed decision about following the automated suggestion. The effectiveness of this decision-making process depends significantly on the type and amount of information provided through these explanations. Recent literature has explored various forms of explanations, with a growing interest in harnessing the capabilities of LLMs for conversational explanations. This involves engaging learners in a multi-step dialogue to enhance their understanding of the recommended content. While the use of chatbots in education is not new \cite{wollny_are_2021}, the use of LLM-powered chatbots for generating learning explanations is still not thoroughly investigated, due to the substantial limitations of LLMs in the critical field of education. Wollny et al. surveyed the tasks that chatbots are used for in education. The authors found that 20\% of the chatbots were used in assisting tasks, while 15\% of the chatbots were used for mentoring. The latter covered three main methods: scaffolding, recommending, and informing. Wollny’s classification intersects with the nine categories found by Yan et al. for the use of LLMs in education: “profiling and labelling, detection, assessment and grading, teaching support, prediction, knowledge representation, feedback, content generation and recommendation.” \cite{yan_practical_2023}. In their survey, the authors also investigated the practical and ethical challenges that face LLMs in education. They point out that the technology readiness level (TRL) of the majority of surveyed papers did not exceed TRL-2. On the ethical front, the majority of the surveyed papers did not reach a transparency tier more than Tier 1 \cite{yan_practical_2023} according to the three tiers of transparency \cite{rodrigo_transparency_2022}, meaning that the proposed approach was only transparent to researchers and practitioners. A transparent approach to educational stakeholders, such as learners and teachers, is not reached by any of the papers the authors surveyed.

This concern is also shared by \cite{fullan_artificial_2023}, where the authors point out several negative effects of ChatGPT in education, such as the lack of originality of its answers that can be meaningless and fail to motivate exploration or imagination by being linear or flat. The authors, however, also emphasize that the use of such technology in education still shows potential, and blocking it is not an option. Therefore, active research is ongoing to mitigate the effects of LLM’s inconsistencies, hallucinations, and bias in education. Among those approaches are model fine-tuning methods and model prompt contextualization. While the former approach requires labeled, domain-specific datasets, the latter approach utilizes the data structures available without the need for labeling. The use of data structures to enrich the LLM’s prompt with additional information that reflects the context of their request is evaluated by \cite{sequeda_benchmark_2023}, who investigate the role of KGs in enhancing the accuracy of LLMs. The authors find that asking the model questions that are posted on a KG representation considerably enhanced the accuracy of the response. We build on this concept, and the potential that KGs offer for modeling educational content, to support an LLM-based chatbot in generating more relevant explanations for a learning recommendation. Our contributions in this paper can be summarized in three points: 1) we propose an approach for using chatbots in recommendation explainability tasks. We use the term “conversational explainability” in this paper and define it as the explainability process that takes place through a bi-directional, multi-step interaction between a user and the system, which happens in a certain use-case and within one context. We consider this type of explanations a natural extension of the conversations between a student and his/her peers or mentors, and thus utilize it in this educational application. 2) We propose a KG-based contextualization approach with experts-in-the-loop, where the context of a GPT-4 prompt is constructed from four categories of information, to enhance the model’s responses. 3) We introduce the role of the chatbot as a potential mediator in a group chat that includes the student, the mentor, and the chatbot. We build an infrastructure to support this use case, which can be extended to peer-group chatting, with the chatbot assistance.

\section{Methodology}
To address the above-mentioned challenges, we design and implement a chatbot module as a part of a web application for learning-path recommendation. The proposed module accommodates the potential of LLMs for providing information and explanations on learning recommendations and provides the student with a channel to connect with human mentors. Four strategies were adopted in this research to achieve the goal of utilizing the capabilities of LLMs and avoiding their susceptibility to errors, bias, and hallucinations:
\begin{enumerate}
\item \textit{We limit the scope of tasks that the LLM-based chatbot is responsible for.} This strategy is meant to use the LLM in tasks where it is less likely to generate irrelevant or wrong information. For example, A user’s question about a general topic, which is not related to the recommendation, is not to be answered by the chatbot in our approach.
\item \textit{We design the dialogue to ensure that the system understands the user’s question.} Re-prompting is utilized in this strategy to confirm the intent of the user. The chatbot will follow up a user question with a statement on how it understood the question. If the user does not confirm the meaning of the question, the chatbot will request rephrasing it, ensuring that the questions lie within the supported tasks. The chatbot will suggest contacting a human if it is not able to understand the user’s question at all.
\item \textit{We enrich the LLM prompt with thorough contextual information.} The context of an LLM request is utilized to guide the text generation towards more relevant content for the user. A prompt’s context may include information about the situation of the user, the previous state of the conversation, additional descriptions of learning materials, domain terminologies, etc. We utilize the KG to extract detailed information on the learning materials and their relations. We also equip the prompt’s context with expert-defined rules regarding the output shape and limits, as well as information from the learning platform and the chat history.
\item \textit{We provide access to mentor support within all the automated tasks.} The connection to a mentor plays two roles in our approach: it allows group chat, in which the student, the mentor, and the chatbot can converse, and it acts as a fallback strategy that the chatbot suggests in the cases where it does not understand a user question, even after re-prompting.
\end{enumerate}

The chatbot lies at the center of the proposed system architecture, see Figure 1. It acts as a focal point between the user, the LLM, and the KG database. At the core of the chatbot is a dialogue manager that controls the flow of information and applies the four strategies mentioned above. To limit the scope of tasks that the chatbot performs, we design an intent classifier, with a pre-defined set of action-intents that are allowed in the conversation. If a user request does not belong to one of the allowed action classes, the system tries to redirect the user to use one of the supported tasks. The following set of tasks are supported in this approach:

\begin{itemize}
\item Asking about the reason behind the recommendation.
\item Asking about the content on the recommendation page in the web application.
\item Asking about the benefits that will be gained from learning a certain learning material in the recommended path.
\item Asking about the relations and similarities between recommended materials and those in the KG.
\item Asking for additional information about the recommended materials.
\item Asking about the relation of recommended materials to the student’s context (e.g., their daily work).
\end{itemize}

\begin{figure}
  \centering
  \includegraphics[width=\linewidth]{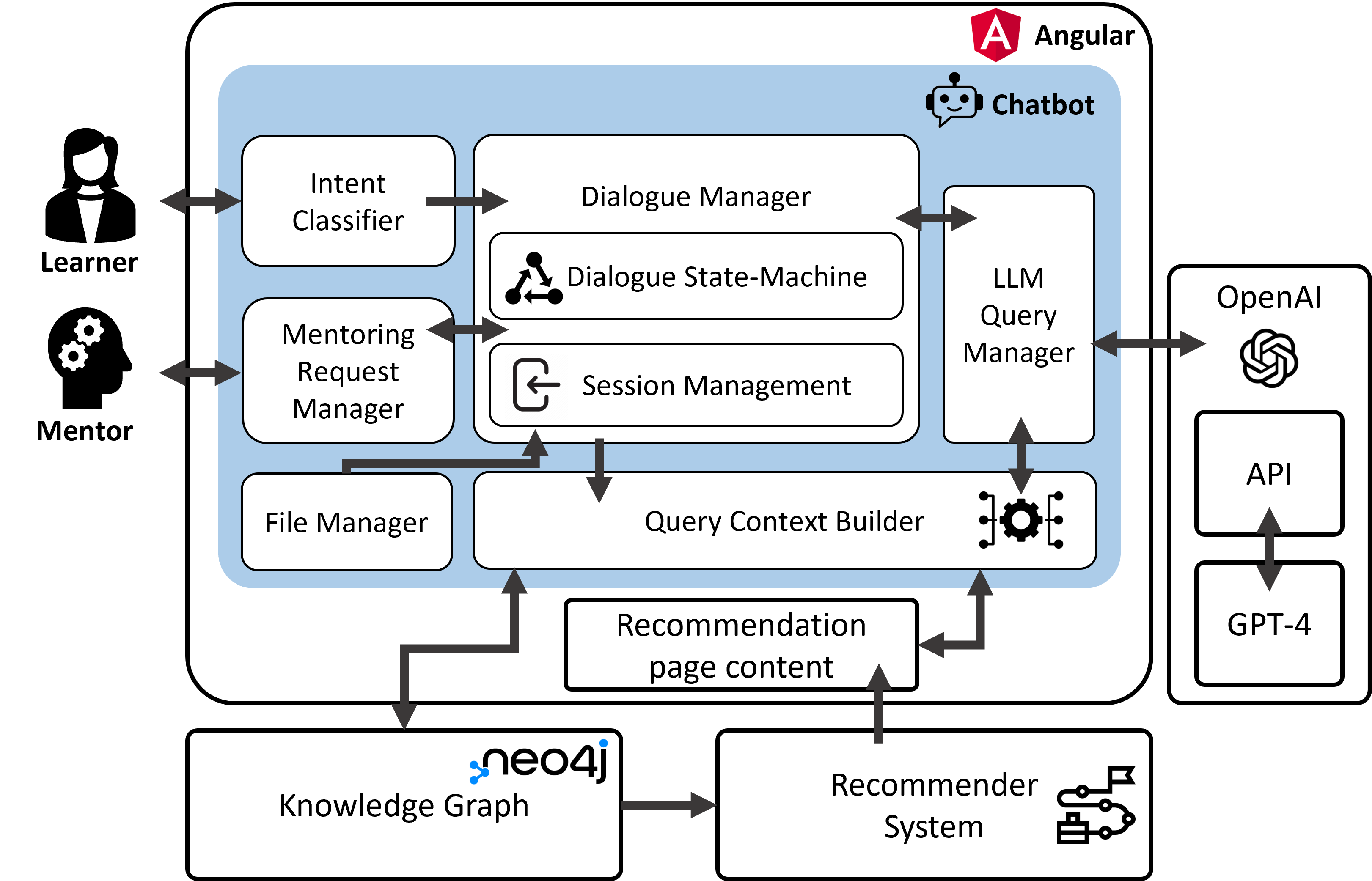}
  \caption{Chatbot system architecture within the Angular Web-App, and its connections to the KG, recommendation algorithm, and the LLM.}
\end{figure}

Any query that does not belong to the previous tasks is classified as “other” and triggers the dialogue manager to request a rephrasing of the query or suggest mentor support. This is accomplished through a state-machine approach, in which we define confirmation states for user queries. The first response to a vague user query goes through a re-prompting state, where the chatbot requests the user to clarify their question. If this does not solve the problem in three re-prompting tries, the dialogue goes to the fallback state, in which the chatbot suggests a connection to a human mentor.

The mentoring-request state is also reachable by a direct request from the student. Mentoring request is initiated from any chat session. The request is then forwarded to available mentors registered in the system. Once a mentor accepts the student’s request, a new session is created by the session manager, in which the mentor, the student, and the chatbot are members, see Figure 2 and Figure 3.

The mentor and student can chat together directly. At the same time, they can ask questions to the chatbot in that session, using the unique identifier (@) in the question, which triggers the chatbot to read that question, call a limited number of previous interactions between the mentor and the student in the chat before the question to use it as contextual information, and then generate an answer. Figure 2 shows the chatbot interface for the tasks of user interaction, mentoring request, and user-mentor chat with the chatbot support. Image- and PDF-file upload feature is supported to utilize the multimodal capabilities of LLMs like GPT-4.

\begin{figure}
  \centering
  \includegraphics[width=\linewidth]{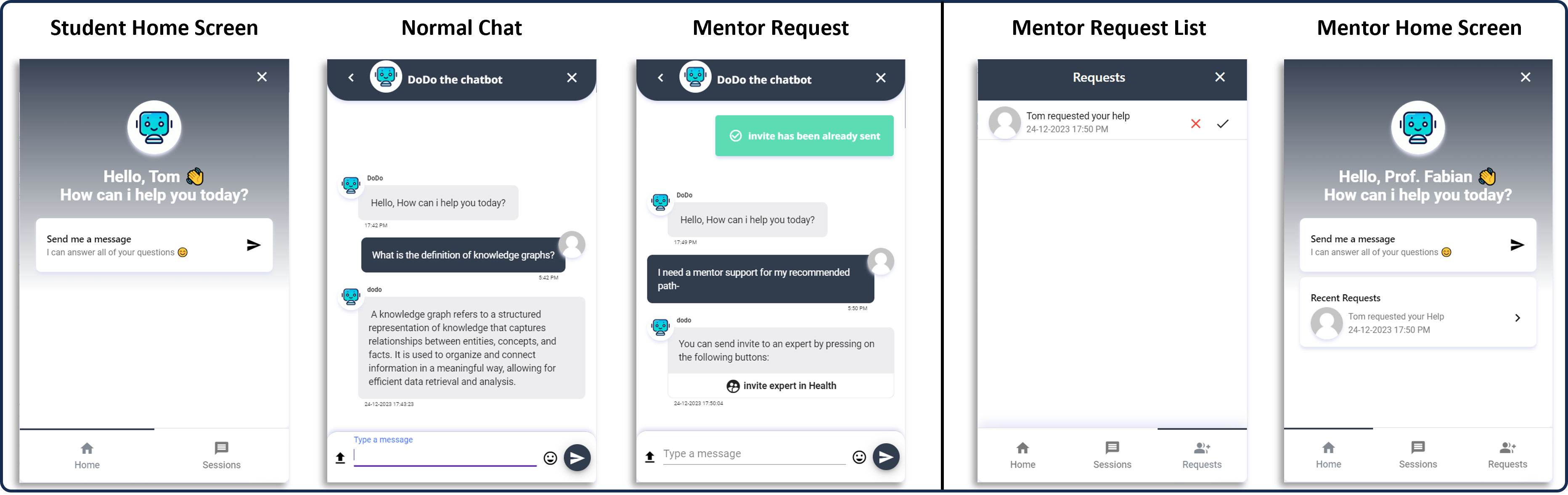}
  \caption{Chatbot interface for user-chatbot and mentor request sessions.}
\end{figure}

\begin{figure}
  \centering
  \includegraphics[width=0.6\textwidth]{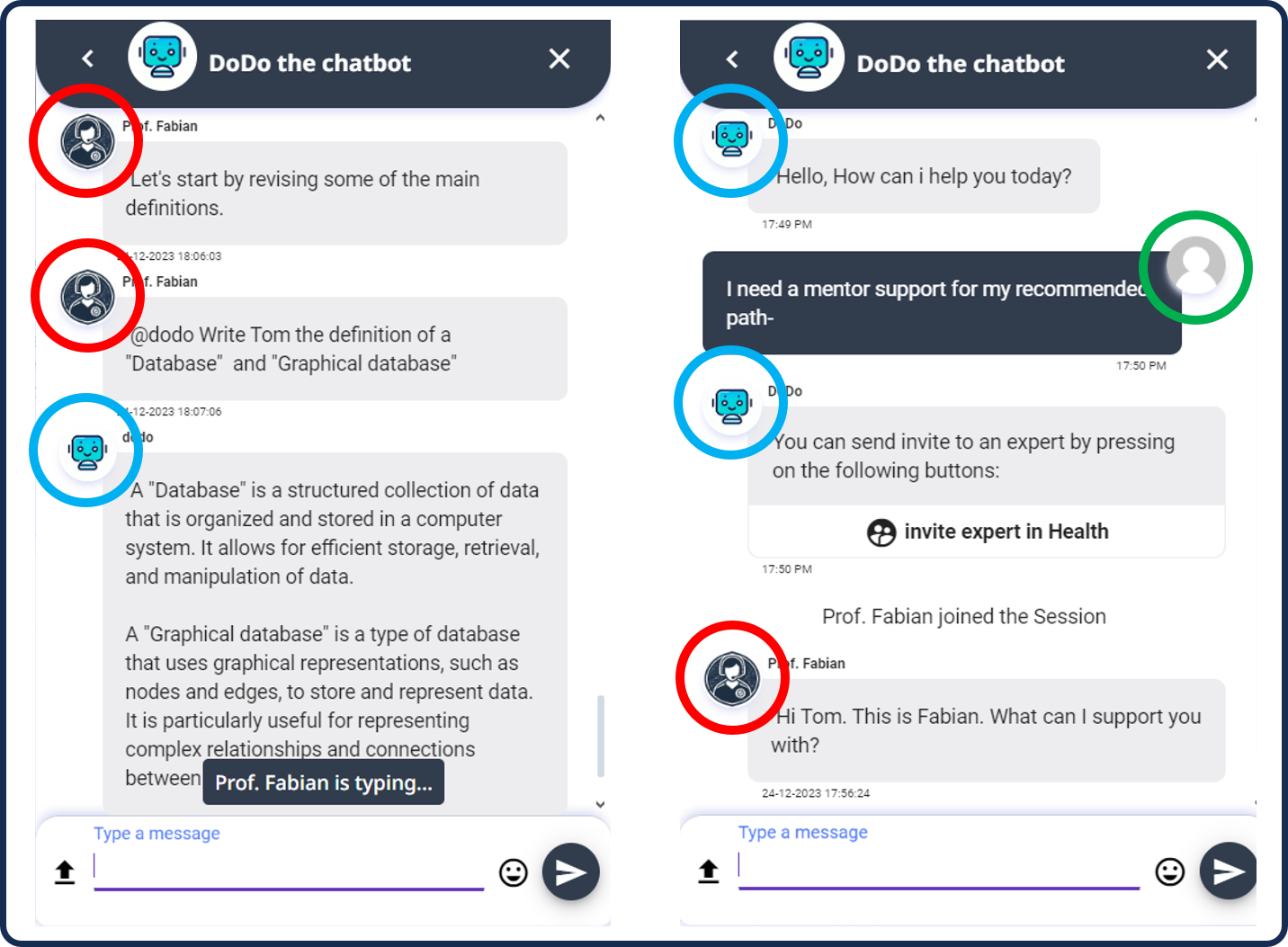}
  \caption{Group chat session.}
\end{figure}

\subsection{Prompt contextualization}
Our third strategy for designing the chatbot is based on the concept of considering the context of a user query to generate a more relevant LLM response. While this concept has shown potential for influencing the output of the LLM, we do not indicate here that more context is necessarily better for the LLM’s output, because it is essential to select the type and volume of contextual information carefully for a positive influence on the output. For the conversational explainability task, the chatbot needs sufficient information about the learning path and its individual elements. It is also important to ensure that the LLM is not generating answers or explanations that a human educator will not accept or use with their students. Therefore, we enroll pedagogy experts and educators in the design phase of prompt contexts, see Figure 4. We survey experts and educators to extract sets of rules that the LLM should follow, such as the length of the response, the type of information to be provided (or not to be provided), and the definitions that the chatbot should consider when generating a response, such as the exact meanings of domain-specific terminology. The integration of domain experts in the design of our system is based on the explainability framework in \cite{abu-rasheed_transferrable_2022}, which describes a set of roles of experts in the KG- and explanation-design processes.

In addition to experts, the other main source of context information in our system is the KG itself. The KG in our case represents a human-curated data structure, which provides comprehensive information on the learning materials and their relationships to each other. Relations among learning materials can form graph communities, in which a learning material is connected to others that appear in the same domain or application area \cite{abu-rasheed_building_2023}. For example, a course on “data analysis” can be created by a computer scientist and thus connected to other similar courses in the domain of computer science, such as “data visualization”, “database management”, and “Python modules for data science”.  The same course can be created by a health expert, and thus connected to another set of courses, such as “patient data privacy”, “digital health records” and “understanding X-ray scans”. Explaining the course “data analysis” is considerably dependent on the context in which it appears. In the KG, course relations can reveal that context, through the other courses connected to it. We use this potential to enrich the LLM prompt’s context with information about the course connections, and the potential KG communities to which it belongs. The KG as a data structure also provides information about the individual courses, the topics they are composed of, and the learning materials in each topic. This includes their metadata, similarity scores to other materials, which relation extraction (RE) algorithms calculated, and the hierarchical connections to the other taxonomical levels, which reflect their curricular format.

Recommended courses and their learning materials are ordered in the learning path by the recommendation algorithm. The path is also a part of the contextual information that the context builder extracts from the RS and the web page on which the recommended path is shown. Including the web page is needed since our Web-App presents the recommendation in multiple ways: textual, structural, and visual. One of the tasks allowed in our chatbot is asking questions about those formats.

To build the prompt’s context from these recourses, we divide the context into four sections: the roles that the chatbot plays, the definitions from the domain, the rules that are to be followed in generating the explanation, and the additional content that is retrieved from the KG.  This context is then added to the main task, or set of tasks, that the dialogue manager defines based on the user’s request. It is noticeable that this arrangement of the final prompt demands a larger volume of text to be transferred to the LLM. This is a compromise that we find necessary to reduce the risk of generating irrelevant outputs, even if the output is not wrong. Depending on the LLM used, the allowed size of context may differ, limiting the amount of contextual text that can be added to the prompt. 

\begin{figure}
  \centering
  \includegraphics[width=\linewidth]{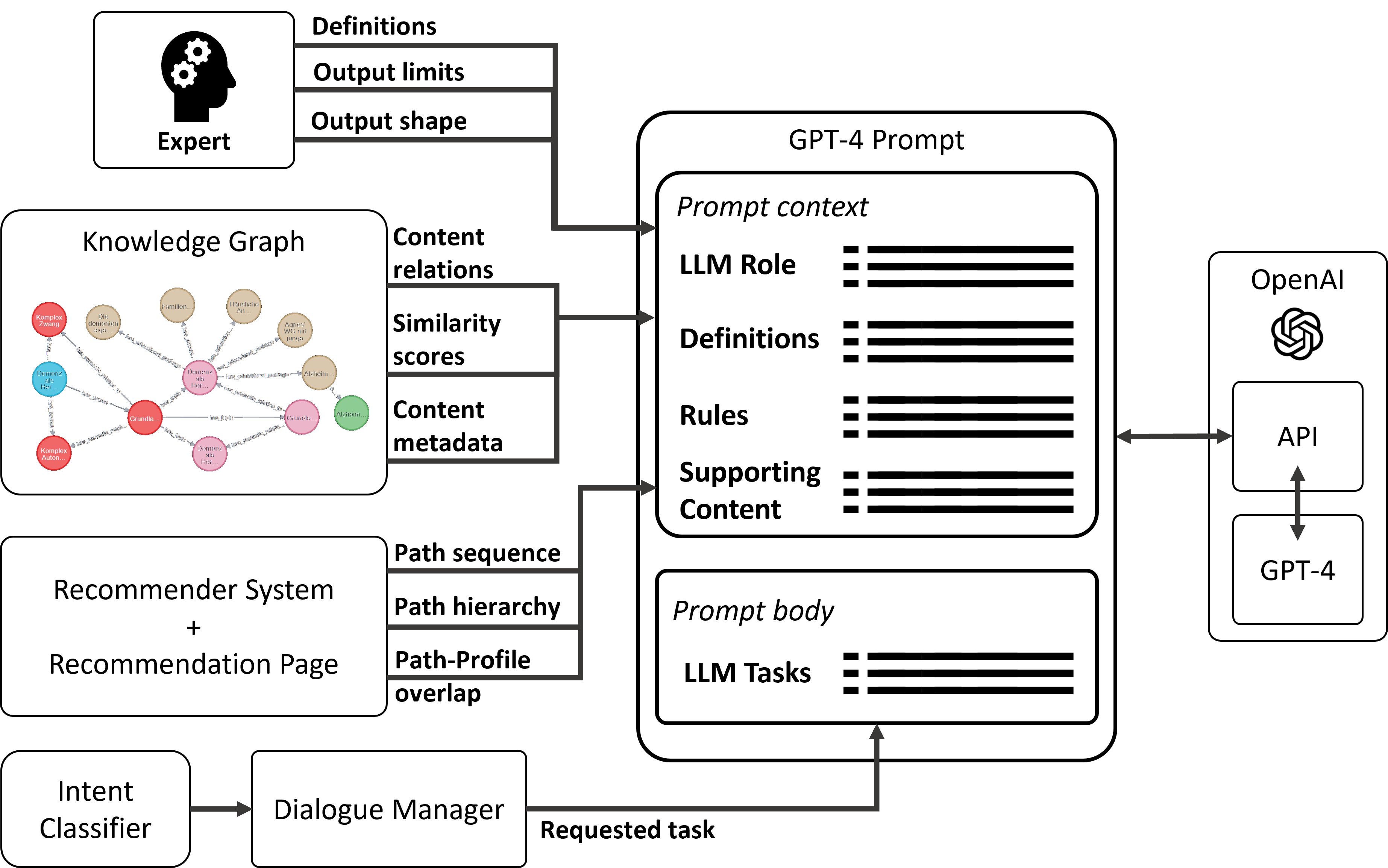}
  \caption{Constructing the prompt’s context from multiple information sources.}
\end{figure}

\section{Proof-of-Concept Evaluation and Results}

To evaluate the system, we test first the performance of the proposed intent classifier. Then, we run a user study to evaluate the features of the explainability chatbot. For the intent classification test, we collect and label a set of 182 user requests, spread equally over the 7 supported intent categories. Figure 5 shows the confusion matrix of the classifier’s performance. The classifier has reached an accuracy level of 88\% over all classes. Table 1 shows the values for the precision, recall, and F1 measure calculated per class. From the performance scores, one can notice that the classifier mostly confuses the questions about the recommendation reasoning with the questions from the third and fourth classes. By analyzing user queries in the test, we notice that several queries about the benefit from a learning material (class 3) and the relations to other learning materials (class 4) were phrased in a way that demands a “justification” of the benefit or the material connections. This may explain the reason why the classifier considered those as questions about the reasoning that justifies the whole recommendation and classified them under class 1.

To evaluate the chatbot features, we conduct a preliminary user study, which is intended as a proof-of-concept, before running a larger-scale evaluation. We design the experiment and survey the test participants to measure their:
\begin{itemize}
\item satisfaction with the chatbot design.
\item satisfaction with the quality of its answers.
\item perception of the correctness of the chatbot answers based on the user intent.
\item satisfaction with the speed of responses.
\item perception of the chatbot’s responses to out-of-scope questions, i.e., in the “other” intent class.
\end{itemize}

For points 2-5, we design a set of eight scenarios, depicted by short stories. Six scenarios require the user to validate each of the intent classes, except class 6 that covers the relation to the user’s daily work. This is because the same recommendation was offered to all users, to avoid introducing a new dependent variable to the experiment through personalization. The seventh scenario validates the chatbot answers in cases it is not able to generate a response. The eighth scenario is meant to validate the mentor-chat session. Likert scales (1-5) and (1-10) were used to record the user answers.

\begin{figure}
  \centering
  \includegraphics[width=0.6\linewidth]{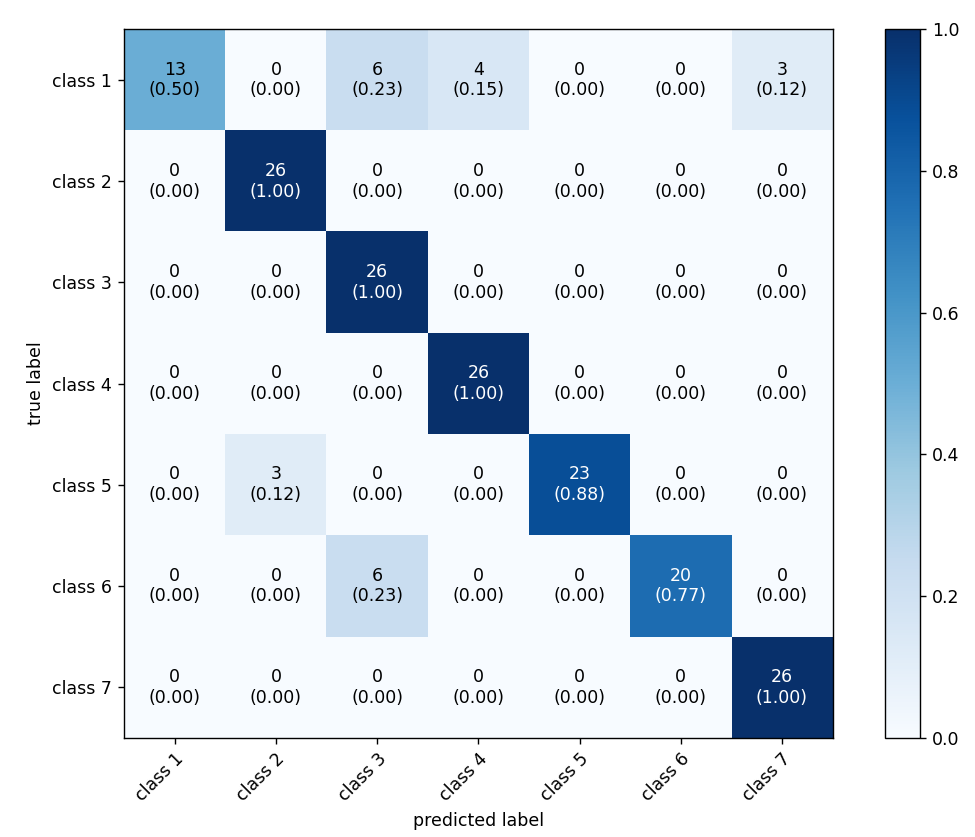}
  \caption{Confusion matrix for the intent classifier.}
\end{figure}

\begin{table*}
  \caption{Precision, Recall, and F1 measure for each intent class}
  \label{tab:eval}
  \begin{adjustbox}{width=\textwidth}
  \begin{tabular}{cllll}
    \toprule
    Class ID &Class description &P & R& F1\\
    \midrule
    1 &Query about the reason behind the recommendation. &0.50 &1 &0.67\\
    2 &Query about the content on the recommendation page in the web application. &0.89 &1 &0.94\\
    3 &Query about the benefits that will be gained from learning a certain learning material in the recommended path. &0.68 &1 &0.81\\
    4 &Query about the relations and similarities between recommended materials and those in the KG. &0.86 &1 &0.92\\
    5 &Query for additional information about the recommended materials. &1 &0.88 &0.94\\
    6 &Query about the relation of recommended materials to the student’s context (e.g., their daily work). &1 &0.77 &0.87\\
    7 &Other queries &0.89 &1 &0.94\\
  \bottomrule
\end{tabular}
\end{adjustbox}
\end{table*}

\begin{figure}
  \centering
  \includegraphics[width=\linewidth]{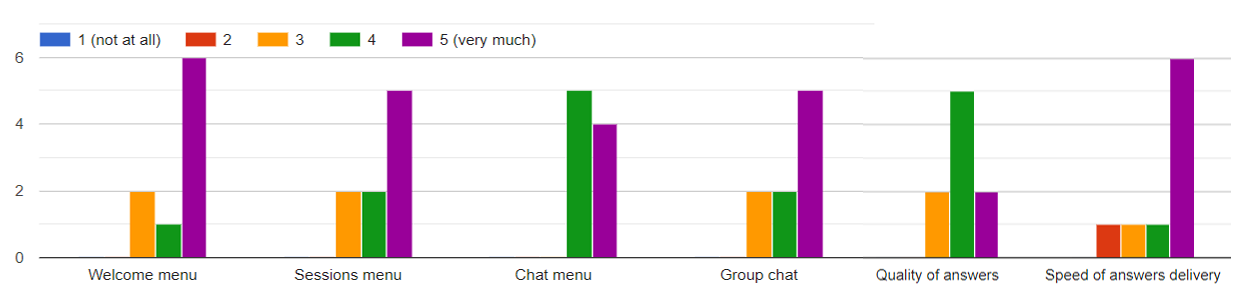}
  \caption{User evaluation of the chatbot menus, group chat, answer quality, and speed.}
\end{figure}

\begin{figure}
  \centering
  \includegraphics[width=0.6\linewidth]{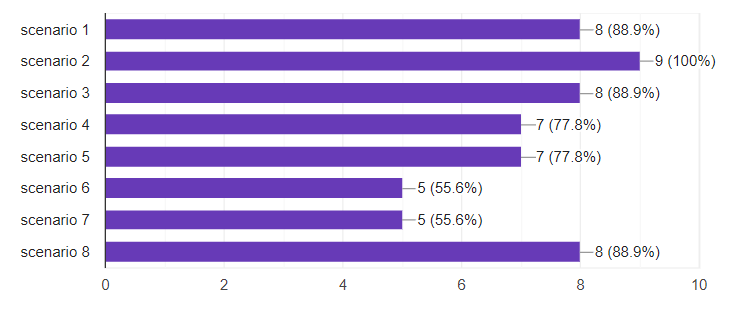}
  \caption{Average user evaluation of the chatbot responses for each of the eight scenarios.}
\end{figure}

We conduct this qualitative experiment with a small sample of nine participants from the target group. The participants were from the academic field and included one post-doctorate, two PhD candidates, and six graduate students. The test duration ranged between 60 and 90 minutes, during which one of the authors took the role of the mentor, due to their knowledge about the recommended path.  
User evaluation of the chatbot’s design and menus, see Figure 6, got an average of 4.7/5. The quality of answers was evaluated with an average of 4.4/5. The speed of response scored an average of 4.6/5. It is important to mention that the speed of responses is dependent on different factors, including the specifications of the servers hosting the chatbot, the speed of the KG search and information retrieval algorithm, and the response time from the API. While we do not have an influence on the API’s response time, we measure the speed of our KG search algorithm and record a response time ranging between 0.72 and 0.9 seconds. The server used for the experiment was equipped with a CPU 13th Gen Intel(R) CoreTM i7-13700KF, 3.40 GHz, and 32.0 GB of RAM. The response time recorded for the complete process, from the user query to the final answer, ranged between four and eight seconds.

Figure 7 shows the results of the eight test scenarios. It can be noticed that the scenarios that cover cases in which the chatbot was allowed to provide an answer received a higher score than the two scenarios (6 and 7), for which the chatbot should avoid providing an answer, but rather re-direct the user to another conversation state, e.g., the fallback state of contacting a mentor.

\section{Discussion and Limitations}

While our evaluation above provides a preliminary proof-of-concept for the proposed approach, one cannot argue that it reflects statistically sufficient evidence that the chatbot has an influence on the learning performance of a student. A larger and more thorough test process is being therefore designed within our ongoing research to test the effect of the chatbot on students, in a longer-term learning setting. This follows the growing concerns, which the authors share, that the rapid development of Gen-AI and the approaches based on it are not accompanied by the same level of real-world testing of the learning effect that those approaches have in real educational settings. From our user study and the involvement of domain experts in the design and evaluation phases of our chatbot, we find evidence for several lessons learned, which we summarize in the following points:

\begin{itemize}
\item The student utilization of the chatbot is greatly influenced by the way they phrase their questions. Even when intent classifiers have high accuracy, two different students may still use very similar sentences and mean different things. LLMs are one solution to support intent classification. However, they require thorough contextualization to understand the sentence’s meaning.
\item LLM outputs are mostly used exactly as the models generate them. However, using parts of the output, or arranging several partial outputs in the pre-defined slots of a larger explanation template offers educators more flexibility for controlling the final explanation content.
\item •	Quantitative evaluation of LLM responses does not necessarily reflect its quality for an education use case. Pedagogy experts pointed out in some of our interviews that several LLM-generated texts are not wrong, but they do not offer a high value-added to students. One expert expressed this idea as: \textit{“[the chatbot response] is not wrong. It is quite fine by me. But I wouldn’t give this answer to my students because it doesn’t enable them to reflect. […] Reflection needs additional information, which is not simply an answer to the question.”}
\item The context added to an LLM query may be phrased in different ways, even when the same information is included. The evaluation of the phrasing’s effect is important, but it presents a complex challenge. In our system, we adjust the context phrasing for each intent, to keep the overall prompt objective -to the best of our ability- and to keep the LLM’s focus on the task itself, without being distracted by the contextual information that may not represent the user intent.
\end{itemize}

The latter point is a limitation of our study since it requires another test for the effect of context phrasing. Other limitations we address for our study include:
\begin{itemize}
    \item The limited sample size of the user study. We consider this evaluation mainly qualitative, which is meant to prove the concept. A larger scale test is being designed within our ongoing research to acquire statistically accepted results, and to focus on the real effect of using the chatbot explanations in learning. For that experiment, an A/B test is designed to compare the effect of conversation explainability against single-step textual explanation modality.
\item In this research, we depend mainly on GPT-4. A comparison is needed to evaluate the results and their potential differences when using other LLMs.
\item Our context does not include user-profile data, which is meant to comply with the GDPR, since we are using an API from a third party. This presented a limitation for the information we use in the context. To solve this issue, a local LLM will be used in the next step to enrich the context with user-specific data from the profiles and study the role of this type of information on the LLM response.

\end{itemize}

\section{Conclusion}
In this paper, we proposed an LLM-supported chatbot approach for conversational explainability of learning recommendations. We focus on harnessing the potential of a GPT-4 LLM while reducing the risks it presents in education. A KG-based design of the LLM prompt’s context was proposed to enrich the LLM’s prompt with thorough information about the context of the student’s query, to enhance its chances for generating relevant and useful output for the student. Our approach is designed to involve educators and domain experts in the design phase of the prompt context and the final shape of the explanation. The chatbot played a mediator role in our system, in which it not only connected the LLM, the KG, and the user query, but also enabled a group chat feature, connecting the student to a human mentor or an experienced peer, to get support in the tasks that the LLM may not perform well. We evaluate the proposed approach quantitatively for validating the intent classification task, and qualitatively through a user study to evaluate the user perception of- and satisfaction with the chatbot’s features and performance. Our preliminary results show a proof-of-concept for the proposed conversational explainability approach and reveal important lessons learned from the design and implementation phases.

%%
%% Define the bibliography file to be used
\bibliography{sample-1col}

\end{document}